\documentclass[onecolumn]{IEEEtran}
\usepackage{cite}
\usepackage{amsmath,amssymb,amsfonts}
\usepackage{graphicx}
\usepackage{textcomp}
\usepackage{xcolor}
\usepackage{graphics}
\usepackage{csquotes}
\usepackage{subcaption}
\usepackage{lipsum}
\usepackage[bookmarks=false]{hyperref}
\usepackage{tabularx}
\usepackage{algorithm}
\usepackage{algpseudocode}
\algblock{Input}{EndInput}
\algnotext{EndInput}
\algblock{Output}{EndOutput}
\algnotext{EndOutput}
\usepackage{tikz}
\usetikzlibrary{matrix}

\begin{document}

\title{A Survey on Blockchain-Based Federated Learning and Data Privacy}

\author{%
Bipin Chhetri, Saroj Gopali, Rukayat Olapojoye, Samin Dehbashi and Akbar Siami Namin\\
\textit{Department of Computer Science, Texas Tech University} \\
\{bipin.chhetri, saroj.gopali, rolapojo, samin.dehbashi, akbar.namin\}@ttu.edu
}

\maketitle

\begin{abstract}

Federated learning is a decentralized machine learning paradigm that allows multiple clients to collaborate by leveraging local computational power and the model's transmission. This method reduces the costs and privacy concerns associated with centralized machine learning methods while ensuring data privacy by distributing training data across heterogeneous devices. On the other hand, federated learning has the drawback of data leakage due to the lack of privacy-preserving mechanisms employed during storage, transfer, and sharing, thus posing significant risks to data owners and suppliers. Blockchain technology has emerged as a promising technology for offering secure data-sharing platforms in federated learning, especially in Industrial Internet of Things (IIoT) settings. This survey aims to compare the performance and security of various data privacy mechanisms adopted in blockchain-based federated learning architectures. We conduct a systematic review of existing literature on secure data-sharing platforms for federated learning provided by blockchain technology, providing an in-depth overview of blockchain-based federated learning, its essential components, and discussing its principles, and potential applications. The primary contribution of this survey paper is to identify critical research questions and propose potential directions for future research in blockchain-based federated learning.

\end{abstract}


\section{Introduction}
\label{sec:intro}

The rapid development of the Industrial Internet of Things (IIoT) has resulted in a significant increase in data generated by connected devices \cite{chanal2020security}. 
The current privacy and security measures for IIoT are outdated and require significant updates. In addition, some of these measures are still under development and testing with a myriad of vulnerabilities. As a result, new techniques and policies are urgently needed to secure data sharing across wireless networks and address security challenges in IIoT. To address these challenges, Monrat et al. \cite{monrat2019survey} proposes the use of blockchain technology as a secure data-sharing architecture and thus introducing the Blockchain technology as a decentralized and secure IoT revolution.
 
Rao et al. \cite{rao2016study} note that user privacy laws in many regions worldwide that mandate technological companies handle user data with extra care. The conventional machine learning techniques have a significant limitation in that they require all data to be gathered in a single location, typically a data center. This approach poses a potential risk to user privacy and could violate data confidentiality laws that protect sensitive information. As a result, new machine-learning techniques that can preserve data privacy and confidentiality are needed to address these concerns. To address the limitations of conventional machine learning techniques, Cloud Service Providers (CSPs) have adopted a strategy of centralizing data storage. This approach helps to ensure data integrity, availability, privacy, and confidentiality. This methodology enables CSPs to manage and protect data more effectively, ensuring that it is secure and readily available to authorized users. By centralizing data storage and management, CSPs can also integrate advanced security measures and technologies to protect against cyber threats and safeguard sensitive information. Nevertheless, CSPs do not always deliver trustworthy data services to customers, and there are problems with cloud data storage such as data breaches, data theft, privacy concerns, and cloud data unavailability. Therefore, submitting raw data to a central server raises privacy and communication concerns for data owners, reducing the likelihood of uploading data.

Bonawitz et al. \cite{bonawitz2019towards} state that Federated Learning (FL) is an innovative approach to machine-learning that resolves problems associated with traditional methods. FL allows multiple parties to train a shared model without revealing their data. The training algorithms on data are distributed across several clients with no need for data samples to be exchanged. A centralized server manages the training process in the traditional FL approach including client management, global model maintenance, and gradient aggregation. The server sends the current model to nodes each round, updated with their data and gradients sent back. The gradients are aggregated and integrated into the model for the next round by the server. FL preserves data privacy by sharing gradient information instead of raw data. However, as Li et al.\cite{li2020federated} note, there is still a risk of sensitive information being exposed to a third party or the central server. Moreover, the conventional FL framework is vulnerable to malicious data and single points of failure, which can undermine its reliability.

Blockchain technology is an alternative to centralized methods in IoT and edge computing, overcoming their limitations. Blockchain's decentralization enables Smart Contracts (SC) to function as a centralized server through blockchain transactions. This study focuses on the implementation of privacy-preserving machine learning methods based on blockchain and their practical applications. The main objective of this exploratory study is to understand how blockchain-based federated learning can enhance the training stage in machine learning algorithms. The focus is on addressing data privacy requirements while enabling practical applications. A review of existing literature on blockchain-based federated learning is conducted. The aim is to provide a comprehensive overview of the current state-of-the-art in this field. The research efforts are categorized and analyzed to identify open research questions and challenges that need to be addressed to advance this line of research. The contributions of this study are as follows:

\begin{itemize}
    \item[--] An in-depth overview of blockchain-based federated learning along with its essential components, underlying principles, and potential applications are presented.
    \item[--] A comprehensive systematic review of existing literature in the context of privacy-preserving in Blockchain is conducted.
    \item[--] A number of critical open research questions and  potential directions for future research in blockchain-based federated learning are proposed.
\end{itemize}

The paper is organized as follows: Section {\bf \ref{sec:motivation}} covers the motivation for the study, while Section {\bf \ref{sec:background}} provides background information on Federated Learning, Blockchain, Blockchain-based Federated Learning, and Data Privacy. Recent studies on implementing Federated Learning using Blockchain are discussed in Section {\bf \ref{sec: recent studies}}. A discussion about the weaknesses of using blockchain technology in federated learning is explored in Sections {\bf \ref{sec:futureWorks}}. The future directions of the research are explored {\bf \ref{sec:futureworks}}. 
The conclusion is presented in Section {\bf \ref{sec:Conclusion}}.

\section{Motivation}
\label{sec:motivation}

The primary motivation behind this study is the pressing issues surrounding data privacy and sharing. Security concerns regarding federated learning have been arisen. These concerns stem from the potential for malicious clients or central servers to attack the global model or access user privacy data.

Qu et al. \cite{qu2020blockchained} tackled the data privacy issue by implementing federated learning, where only the models and not the raw data are shared. This ensures the data's efficiency and usefulness while maintaining privacy. A fully decentralized federated learning system is proposed. The system employs blockchain technology as the underlying architecture and the proof-of-work (PoW) consensus process. The 
decentralized system offers resistance to poisoning attacks. Incentives are provided and accuracy is enhanced through member selections.

Data leakage during storage, transmission, and sharing is a significant challenge faced by data owners and providers. This challenge is particularly prominent in the context of Industrial Internet of Things (IIoT) applications \cite{sisinni2018industrial}. To address attacks on global models or user privacy data, Li et al. \cite{saroj1} proposed the Blockchain-based Federated Learning framework with Committee consensus (BFLC), a decentralized, blockchain-based federated learning framework. However, the Committee consensus mechanism (CCM) used in the framework may result in a large amount of communication overhead between nodes. This could lead to slow training times and increased energy consumption.

Lu et al. \cite{saroj2} proposed a solution to data-sharing challenges by combining federated learning with permissioned blockchain. The permissioned blockchain in this system creates secure connections between end IoT devices. These connections are established through encrypted records maintained by supernodes, such as base stations and roadside units. This ensures data privacy and accessibility. The proposed architecture does not store raw data. It uses permissioned blockchain to access related data and controls data accessibility. This addresses storage constraints and privacy concerns.

To facilitate collaborative learning and data protection, it is crucial to conduct research in blockchain-based federated learning and data privacy. In conventional machine learning scenarios, sharing a large dataset required for training a model is challenging. This is particularly true when the data is distributed across multiple organizations or individuals \cite{konevcny2016federated}. Federated learning solves the problem of sharing a large dataset for training a model. It does so by allowing each party to train a local model on their data. The local model's parameters are then shared with a central server, which aggregates these updates to produce a global model \cite{mcmahanmooreramagehampson17}.

To enhance the security and trustworthiness of this process. Blockchain technology can provide a tamper-proof and transparent ledger for recording the model updates \cite{ateniesekamarakatz15}. This way, participating parties can verify the integrity of the global model. This ensures equitable incorporation of their contributions. Furthermore, blockchain can be used to implement privacy-preserving mechanisms. These mechanisms, such as differential privacy, enable participating parties to share their model updates. The sharing of updates is accomplished without revealing sensitive information.

\section{Technical Background}
\label{sec:background}

This section provides a discussion of the background of federated learning, blockchain technology, blockchain-based federated learning, and data privacy along with the fundamental concepts and principles of each area. 
Understanding the backgrounds of blockchain and federated learning is crucial. It allows us to appreciate the potential benefits and limitations of applying blockchain-based federated learning. This application can address data privacy concerns in collaborative learning scenarios.

\subsection{Federated Learning}

Federated Learning (FL) is a decentralized approach to machine learning. FL allows clients to participate in training a machine learning model. Clients can contribute to training without uploading their data samples to a centralized data warehouse. This preserves the privacy of each data sample, as they remain with the clients. In FL, clients train a model locally on their private data samples. This local training happens in each iteration of model training. Clients use an initial global model provided by the aggregation server. After training, the model gradients or weights are generated and uploaded to a server for aggregation. The core workflow involved in traditional FL includes several steps. The first step is the selection of participating clients.
Clients then train their own local model. The local model weights are uploaded to the centralized server. Finally, the server aggregates the local model weights to obtain a global model \cite{R1}.

\subsection{Blockchain}

Blockchain is a decentralized and tamper-proof system used for permanently recording transactions. It consists of a network of nodes, transactions, a chain of blocks, and a shared ledger. The transactions are recorded and maintained by all the nodes in the network, providing benefits such as decentralization, immutability, transparency, and anonymity \cite{R2}. Based on the level of access control, blockchain can be categorized as public, private, or consortium. In a public blockchain, any node can participate in the network without requiring permission. In contrast, nodes on a private blockchain require authorization to join the network and access the shared ledger. In a consortium blockchain, control is typically restricted to selected nodes. These selected nodes have the right to generate new blocks. This makes consortium blockchain a partially decentralized system \cite{R1}.

\subsection{Blockchain-based Federated Learning}

The traditional approach to Federated Learning (FL) depends on a central server for model aggregation, which can be considered as a weak point in the system. To improve reliability, blockchain technology, which is decentralized, has been proposed as a solution \cite{R3}. Blockchain-based Federated Learning (BCFL) integrates the decentralized property of blockchain with the distributed nature of FL to eliminate the threat of a single point of failure in the FL system's aggregation server \cite{saroj1, R4}. Various BCFL architectures have been proposed, which can be categorized into three groups: (1) fully coupled BCFL, (2) flexible coupled BCFL, and (3) loosely coupled BCFL 
\cite{R1}.

Smart Contracts (SC) have been used to implement the functions of FL aggregation in more recent approaches such as BLOCKFL \cite{R4} and BAFFLE \cite{R10}. These SCs are activated through blockchain transactions to facilitate the aggregation of local model updates from participating clients. In BLOCKFL, a two-phase consensus protocol is used to ensure the integrity of the aggregated model. BAFFLE employs a modified Byzantine fault-tolerant consensus algorithm. The algorithm is used to address the security and scalability issues faced in traditional FL systems \cite{R10}.



\subsection{Data Privacy}

In traditional FL, the data samples of each participating client are not exposed to each other or to the aggregation server. However, the local model updates sent over for aggregation are exposed to the server. Typically in a BCFL, the model updates from the participating clients are also uploaded to the blockchain network as raw data. These scenarios lead to data leakage and pose a threat to the system as malicious clients or attackers could exploit this vulnerability \cite{saroj2}, \cite{R11}, \cite{R12}. Li et al. \cite{R14} proposed a blockchain-based collaborative system. The system, called BLADE-FL, is designed to share data across distributed multiple parties. The goal is to reduce the risk of data leakage. They ensured data privacy by incorporating differential privacy into federated learning. Other techniques, such as Homomorphic Encryption \cite{R15},  and Secure Multiparty Computation \cite{R16}, have been integrated to preserve privacy from end to end in a BCFL. These data privacy techniques are briefly discussed as follows:

\subsubsection{Differential Privacy (DP)}
A mathematical definition of privacy that protects users' privacy in published data by adding some randomly generated noises. Basically, the definition of DP is stated as for a random algorithm $A$, $Q$ is the set of all possible outputs. If for any pair of the neighboring datasets $D$ and $D'$, any subset $S$ of $Q$, algorithm $A$ satisfies.
The algorithm $A$ satisfies differential privacy, where $\varepsilon$ is the privacy protection budget.
\begin{equation}
    \operatorname{Pr}\left[A(D) \in S\right] \leq e^{\varepsilon} \operatorname{Pr}\left[A\left(D^{\prime}\right) \in S\right]
\end{equation}

The above equation represents a probabilistic inequality. The inequality indicates that the probability of algorithm $A$ producing a result in set $S$ with dataset $D$ is less than or equal to $e$ raised to the power of $\varepsilon$ times the probability of obtaining a result in set $S$ with a different dataset $D'$. In simpler terms, using dataset $D'$ increases the likelihood of obtaining a result in set $S$ compared to dataset $D$, by a factor of $e$ raised to the power of $\varepsilon$.

\subsubsection{Homomorphic Encryption (HE)}

This is a cryptography method that allows computation on encrypted data and provides encrypted results to the user without having to decrypt the encrypted data. Asides from being able to process encrypted data, HE also ensures that the privacy of data is preserved. Given encryption of messages ($m_1, m_2, m_3,....m_n$) as $E(m_1), E(m_2), E(m_3)...., E(m_n)$, a ciphertext that efficiently encrypts $f(m_1, m_2, m_3....m_n)$ can be computed for any  computable function $f$.

\subsubsection{Secure Multiparty Computation (MPC)}
Is also a cryptography technique that enables different parties to carry out distributed tasks in a well-secure manner. In an MPC, a given number of parties, {$ P_1, P_2, P_3,...., P_N$} and each have private data, {$ D_1, D_2, D_3,...., D_N$} respectively. The participants individually compute the value of a public function on their private data as; $F(D_1, D_2, D_3,...., D_N)$ while keeping their individual inputs secret.

\section{Blockchain-based Federated Learning}
\label{sec: recent studies}

This section reviews existing research studies related to the application of blockchain technology to federated learning. 

\subsection{Homomorphic encryption based}

Wang et al. \cite{wang2022blockchain} proposed a blockchain-based privacy-preserving federated learning (BPFL) model in the Internet of Vehicles (IoV). The main goal is to mitigate the privacy risk of poisoning attacks by participants and stealing sensitive data by aggregation servers.  BPFL model consists of four sections:  Client user, federated learning node (FL node), model aggregation node (MA node), virtual verification node (VV node), and certificate authority (CA). Using homomorphic encryption and developing Multi-Krum\cite{blanchard2017machine} lead to verifying and filtering local model changes. Multi-Krum uses the Krum function to calculate scores for each proposed vector, which helps identify reliable participants while excluding outliers in distributed machine learning. As a result, the system decreases runtime overhead \cite{wang2022blockchain}.

In another study, Miao et al. \cite{miao2022privacy} provided privacy by designing a blockchain-based privacy-preserving byzantine-robust federated learning (PBFL) model. They make a trusted global model by checking cosine similarities to identify negative gradient and honest gradient vectors. Also, they applied Cheon-Kim-Kim-Song (CKKS) scheme, a fully homomorphic encryption (FHE) method to encrypt local gradients and provide privacy protection. However, their work is suitable for a balanced distribution of client data only and not in cases where the client data is Non-independent and identically distributed (Non-IID). 

Sun et al. \cite{sun2022blockchain} approach to blockchain-based federated learning involves encrypting gradients using the BCP (Bresson-Catalano-Pointcheval) mechanism, which adds noise to each encrypted gradient. Then, all the updated gradients are collected to another blockchain. This blockchain could evaluate the malicious client if they provided a low-quality gradient. In terms of overhead, the algorithm does not add any extra overhead of encryption compared to previous works, but it reduces the time-consuming process. Also, the accuracy of the audit algorithm in the baseline is over 92\%, and it decreased to 90\% when faced with a poisoning attack (i.e., low gradients). This shows that the proposed model can recognize malicious owners.  However, as the number of data owners increases, the behavior and audit chains may become overwhelming. This can lead to increased processing times and delays. This could limit the practical use of the proposed approach in large-scale federated learning scenarios.

A novel technique was proposed by Alzubi et al. \cite{R17} using deep learning and blockchain paradigms to preserve the privacy of electronic health records. First, the health records are classified using CNN into normal and abnormal users. Next,  a federated learning mechanism based on cryptography is integrated into the Blockchain system. The blockchain becomes responsible for keeping track of encrypted local models from the FL clients. The blockchain also ensures that the client's contributions to the global model are verified before aggregation.

An efficient and secure blockchain-based FL system paradigm (ESB-FL) was also developed by Chen et al.\ \cite{R19}. In the introduced scheme, a new lightweight cryptography tool was proposed. The tool is based on a non-interaction designated decryptor function. The tool is used to encrypt each participant's local model updates. ESB-FL can ensure the privacy protection of FL participants. ESB-FL can also efficiently preserve the global model's accuracy. The approach achieves this with considerably low and effective communication costs.

\subsection{Differential Privacy-based Approaches}

Zhao et al. \cite{zhao2020privacy} designed a blockchain-based federated learning model for home appliance manufacturers to develop their services and products. First, customers train a model using a collection of home appliance data. Then, they send the trained model to the blockchain to trace clients' or manufacturers' activities and prevent the probability of cyber threats. Finally, as a miner, one of the clients uploads the model to the blockchain. The authors recommended using differential privacy techniques on features to provide clients' privacy by adding $\varepsilon$-DP noise to features. 

In another study, Qi et al. \cite{qi2021privacy} proposed a federated learning-based Traffic Flow Prediction (TFP) system. They have integrated GRU neural networks with blockchain and FL-based TFP schemes. Rather than directly sending individual data, the participating vehicles use their data to perform local model training and share local model updates, thus protecting privacy. Blockchain prevents the security risks of the central server and clients. This is achieved by replacing the central server with a group of trusted nodes. The nodes manage all the local model updates. In addition, they apply differential privacy by adding Gaussian noise to local model updates, thereby protecting the client’s data. The proposed model was compared with LSTM, stacked autoencoder (SAE), and SVM, in which the proposed model could accurately predict traffic flow better than the other models. Moreover, the proposed model effectively mitigates poisoning attacks since the accuracy of blockchain does not reduce even if the number of malicious clients increases. The proposed model faces a challenge in that the convergence rate of the SAE model is faster. This is because the SAE model does not need to complete a model aggregation step. The SAE model has a centralized learning paradigm, which contributes to its faster convergence rate.

Wan et al. \cite{wan2022privacy} proposed a novel blockchain-based federated learning framework to avoid data falsification beyond 5G networks (B5G) enabling edge computing. They also added a differential privacy identifier to Wasserstein Generative Adversarial Network (WGAN)\cite{arjovsky2017wasserstein} to distinguish if synthetic data complies with differential privacy. Lastly, a time delay analysis was conducted on a single epoch of the proposed model, which was then used to determine the optimal rate for generating blockchain blocks. The trained local parameters of edge devices are regenerated by the WGAN generator and then assessed by DP-identifier and WGAN discriminator during the FL training process. With better data utility, this technique ensures that the resulting synthetic model parameters fulfill DP. Blockchain-enabled FL's convergence latency has been seen to be quadratic to the block production rate. As a result, the experimental findings lead to an optimum block generation rate. 

Shayan et al. \cite{shayan2020biscotti} introduced Biscotti, a blockchain-based system for federated learning. It uses cryptography and blockchain technology to enable secure and private federated learning across multiple organizations. The system allows organizations to store and process data locally. The system also allows machine learning models to be trained across all participating organizations. Biscotti comprises four main components: blockchain ledger, consensus protocol, smart contracts, and off-chain storage. The system provides a variety of security measures such as data privacy and access control. The measures are put in place to ensure that data is secure and only accessible to authorized parties. Additionally, the system utilizes various techniques to facilitate efficient and secure data exchange, such as differential privacy and distributed data aggregation.

Salim et al. \cite{R18} proposed a differential privacy blockchain-based explainable FL (DP-BFL) architecture using Social Media 3.0 networks. This architecture allows internet-enabled devices to participate in training global models while preserving data privacy. After local training,  participants upload their deferentially private local model updates to the blockchain system. These local updates are then evaluated and verified by the miners of the blockchain system. DP-BFL ensures that the privacy of the participants as well as a good performance of the global model, is achieved by mitigating the impact of malicious participants' local updates.

Qu et al. \cite{9839634} proposed a novel approach to block-chain-enabled adaptive asynchronous federated learning (FedTwin). The approach enables adaptive and asynchronous training in digital twin networks. The approach addresses the challenges of centralized processing, data falsification, privacy leakage, and lack of incentive mechanisms in digital twin networks. FedTwin uses a proof-of-federalism consensus algorithm for efficient and secure synchronization of digital twin networks (DTN), enabling a personalized incentive mechanism. The approach also uses privacy-preserving local digital twin (DT) training with falsification filtering. The approach uses adaptive asynchronous global aggregation of DTN with a roll-back mechanism. The authors evaluate the performance of FedTwin on a real-world dataset, which shows its superior performance for DTN.

\subsection{Secure Multi-party Computation-based Approaches}

Lu et al. \cite{saroj2} proposed a collaborative architecture enabled by blockchain to share data among multiple parties. The architecture also minimizes the risk of data leakage and grants data owners greater control over access to shared data. By using federated learning to construct data models and share them instead of raw data, the authors transformed data sharing into a machine learning problem, thereby, enhancing the usage of computing resources and the effectiveness of the data-sharing system. To safeguard data privacy, the authors integrated differential privacy into federated learning. The effectiveness of the proposed model was evaluated for data categorization using benchmark, open real-world datasets. However, three potential threats were identified: data quality, data security, and data authority management. To address these threats, the authors integrated federated learning to achieve differential privacy. They employed a permissioned blockchain to eliminate centralized trust. They ensured the quality of shared data to prevent invalid sharing. They facilitated secure data management by allowing data providers to upload data only through permissioned blockchain.

Li et al. \cite{saroj1} proposed BFLC, a Blockchain-based Federated Learning framework with Committee consensus. To address the integrating storage optimization, analysis of hostile node threats, and community node administration issues, FL is performed by participating nodes using blockchain. The blockchain maintains global models and local updates without the use of a centralized server. The authors employ a novel delegated consensus mechanism, which addresses the missions of gradient selection and block generation while accounting for the communication cost of FL. In the experiment, BFLC demonstrated higher accuracy compared to basis FL and stand-alone framework \cite{Saroj1.1} in various malicious proportion settings. The author incorporated real-world datasets into the BFLC framework, enabling them to obtain global models that closely resemble the centralized training approach in federated learning. However, the need for a trusted blockchain system raises unexplored aspects of ensuring trustworthiness, which may pose challenges and require further investigation in order to address potential vulnerabilities and maintain the integrity of the BFLC framework.

Qu et al. \cite{qu2020blockchained} proposed a decentralized paradigm for big data-driven cognitive computing (D2C). This paradigm addresses issues such as unreliable performance, inefficiency, privacy leakage, and poisoning attacks. It does so by combining federated learning and blockchain. Their novel architecture uses the federated learning paradigm for massive D2C. This architecture significantly improves the manufacturing performance of Industry 4.0 \cite{lasi2014industry}. It also overcomes privacy and performance problems associated with cognitive computing. To enhance performance, accuracy, and incentive mechanisms for Industry 4.0 automation, the authors integrate blockchain into federated learning, creating a D2C paradigm for the Industry 4.0 model. They develop an optimization model with a modified Markovian decision process to simulate a conflict with adversaries. The model increases accuracy and robustness against poisoning attacks.



Rehman et al. \cite{R8} presented a novel approach for secure and privacy-preserving federated learning.
The proposed approach is based on blockchain technology, providing a distributed consensus mechanism for reputation-aware federated learning. 
This system allows the federated learning participants to securely and privately exchange information. It also ensures data privacy and integrity. The system uses a blockchain-based distributed ledger to provide a trustless and decentralized environment for federated learning. 

 Arachchige et al. \cite{arachchige2020trustworthy} proposed PriMod-Chain, a new framework for trustworthy and privacy-preserving machine learning in Industrial IoT systems. To ensure privacy and trustworthiness, the framework combines smart contracts, blockchain, Federated Machine Learning (FedML), Differential Privacy (DP), and InterPlanetary File System (IPFS). The proposed framework was tested for feasibility, and the results for privacy, security, reliability, safety, and resilience were all positive. The authors suggested further research to reduce latency in order to improve efficiency. The PriMod-Chain protocol is presented as a viable solution for reliable privacy-preserving machine learning in Industrial IoT systems.

\begin{table*}
    \caption{Overview of recent studies on Data Privacy in BCFL.}
    \label{tab1}
    \centering
    \begin{tabularx}{\linewidth}{|X|X|X|}

            \textbf{Paper} & \textbf{Privacy Approach} & \textbf{Challenges}  \\ \hline
            Wang et al. \cite{wang2022blockchain} & BPFL- A combination of Multi-Krum and homomorphic encryption & Efficient model combination and complex homomorphic encryption management. \\ \hline
            Zhoa et al. \cite{zhao2020privacy} & DP- Laplace noise & Optimizing noise level and selecting privacy parameters. \\ \hline
            Miao et al. \cite{miao2022privacy} & PBFL- Fully homomorphic encryption and cosine similarity & Scheme on a balanced distribution of client data and not on cases where the client data is non-IID \\ \hline
            Sun et al. \cite{sun2022blockchain} & Homomorphic encryption (BCP-based) for gradient & Blockchain-based audit approach for encrypted gradients may have limited scalability due to the increased processing times and delay \\ \hline
            Li et al. \cite{saroj1} & BFLC- Blockchain-based Federated Learning framework with Committee consensus & CCM approach has increased energy consumption due to a large amount of communication overhead involved during model updates between nodes \\ \hline
            Lu et al. \cite{saroj2} & PBFL- Privacy-preserving data sharing the mechanism for distributed multiple parties & Improving the utility of data models mapped from raw data is necessary \\ \hline
            Qu et al. \cite{qu2020blockchained} & BFL- Decentralized paradigm for big data-driven cognitive computing (D2C) & Improves on Markov decision process (MDP) rather than addressing privacy issues with blockchain that is assumed tamper-proof \\ \hline
            Wan et al. \cite{wan2022privacy} & BFL- Wasserstein generative adversarial network (WGAN) & Need for efficient communication and computation methods, privacy and security concerns in federated learning, and the problem of non-iid data distribution in edge computing environments \\ \hline
            Shayan et al. \cite{shayan2020biscotti} & Biscotti: a fully decentralized peer-to-peer (P2P) approach to multi-party ML & Requiring large honest samples for Multi-Krum, limited scalability for large deep learning models, and vulnerability to privacy attacks. \\ \hline
            Qi et al. \cite{qi2021privacy} & BFL- Traffic Flow Prediction (TFP) & Slower convergence rate due to its decentralized learning paradigm and model aggregation step, as opposed to the SAE model's centralized learning. \\ \hline
            Ur Rehman et al. \cite{R8} & BFL- Reputation-aware fine-grained & A reputation-aware federated learning system that exchanges information securely and privately while maintaining data privacy and integrity. \\ \hline
            Arachchige et al. \cite{arachchige2020trustworthy} & PriModChain - Differential privacy, Federated ML, Ethereum blockchain, and Smart contracts. & Vulnerabilities and providing security recommendations. \\ \hline
            Wang et al. \cite {wang2020ai} &BEMA -   Multiparty multiclass margin  System initialization, 
            off-chain sample mining and on-chain mining. & Lack of guaranteed robustness against Byzantine attacks. \\ \hline 
            Alzubi et al. \cite{R17} & Deep learning and Blockchain techniques for electronic health record privacy-preservation & User classification, integration using cryptography, and client contribution verification prior to model aggregation.  \\ \hline
            Salim et al. \cite{R18} & DP-BFL - Differential Privacy blockchain-based explainable FL & Ensuring participant privacy, maintaining global model performance, and mitigating the impact of malicious local updates. \\ \hline
            Chen et al. \cite{R19} & ESB-FL - Blockchain-based FL system paradigm using Cryptography & To protect FL participants' privacy, maintain global model accuracy with low communication costs. \\ \hline
            Liu et al. \cite{R20} & Privacy-Preserving permissioned Blockchain enabled FL with Multi-Party Computation and Fully Homomorphic Encryption & Privacy protection of participants, anonymity, and secure model updates using multi-party computation and fully-homomorphic encryption. \\ \hline
            Qu et al. \cite{9839634} & BFL - Digital twin networks (DTN) & Challenges in digital twin networks include centralized processing, data falsification, privacy leakage, and lack of incentive mechanisms. \\ \hline
    \end{tabularx}
\end{table*}

Wang et al. \cite{wang2020ai} introduced a blockchain-empowered decentralized, secure multiparty learning system called BEMA, where learning parties hold diverse local models. Their work suggests "on-chain" and "off-chain" mining strategies for defense against attacks. The proposed approach involves two steps. The first step is to identify data samples suitable for model calibration. On the other hand, the second step is used to calibrate particular local models based on the discovered samples. Then, these models are entered into the new blocks. Mainly, BEMA includes off-chain and system initialization, both on-chain mining and sample mining. During system startup, The operators (OPs) register their names (IDs) and model details on the link. The participating party can then register their IDs and chain model information. Once a miner uses a valid sample to update models on the chain, each party can broadcast it to the blockchain and earn certain system rewards.

Lu et al. \cite{R20} proposed a blockchain-enabled secure federated learning system for distributed banks. This system combines multi-party computation (MPC) and the multi-key fully-homomorphic encryption(FHE) scheme. The local model updates from the participants are encrypted using the Multi-key FHE scheme and then signed with some pseudo-ID before sharing with others participating in the MPC. With this approach, the participants' privacy protection and anonymity is conveniently achieved.

\section{Data Privacy Challenges}
\label{sec:futureWorks}

Several studies have proposed different solutions to implement federated learning using blockchain technology. Table \ref{tab1} lists these studies, which aim to address privacy challenges in federated learning. However, the approaches employed by these studies encounter some challenges such as efficient model combination, selecting privacy parameters, non-IID data distribution, scalability, and privacy attacks. Some proposed solutions include homomorphic encryption, differential privacy, reputation-aware federated learning, and digital twin networks. These approaches aim to maintain global model accuracy, protect participant privacy, and reduce the impact of malicious local updates. However, they also have limitations, such as increased energy consumption, communication overhead, and security vulnerabilities. Despite these challenges, data privacy and security concerns still arise. Although blockchain technology has effectively decentralized federated learning, it has some drawbacks. 

\subsection{Data Leakage} 
The privacy of BCFL can be compromised by inference attacks even though the model updates are encrypted. Malicious users can still analyze the updates to deduce information. To address this issue, future works may explore the use of other FL structures or combine different techniques.

\subsection{Model accuracy and Latency}
BCFL models require maintaining and improving the accuracy and efficiency of the model. The mini-batching at the client during each training epoch and increasing multi-client parallelism to reach a target test-set accuracy framework can be adopted.
The learning performance of Federated Learning has not been discussed in details. It is necessary to verify the multi-key encryption protocol because of the way it secures the federated ML model data. The accuracy and latency of PrimodChain \cite {arachchige2020trustworthy} systems still need optimization. 

\subsection{Unexplored Complexity}
The work presented by Wang et al. \cite{wang2020ai} provides only theoretical analysis, and BEMA robustness against Byzantine attacks cannot be guaranteed. A Byzantine attack occurs when an attacker adheres to the system protocol but disseminates malicious information to innocent system participants, with the goal of diminishing system performance and manipulating or influencing the system's output. More evaluation is still needed as the learning strategies and security concerns are not fully investigated. The existing research on multiparty learning has mainly focused on homogeneous local models. However, there is still a lack of research on multiparty learning over heterogeneous local models. This is despite the fact that such a scenario may be more practical and useful in real-world applications.

\subsection{Incentive Mechanism Scheme}
The blockchain incentive mechanism plays a crucial role in motivating users to participate in consensus. Abandoning the token incentive could significantly decrease users' motivation to participate if the incentive system of the agreement is not sufficient or perfect enough. The rewards are typically accessible when a new block is
either automatically generated or obtained by charging fees for transactions. Blockchain technology requires honest mining for the successful completion of mining the block. The significant role of incentive mechanisms schemes in blockchain federated learning is often overlooked in many studies. 

\subsection{Scalability}

The blockchain-based audit approach for encrypted gradients in federated learning provides privacy while assessing gradient quality. However, scalability can be a challenge due to the consensus requirement in blockchain technology. Adding new blocks necessitates agreement among all network nodes, resulting in time-consuming and expensive audits, particularly for large federated learning systems. Two approaches to mitigate scalability limitations are (1) off-chain computation \cite{eberhardt2018off}, which performs audits on a subset of nodes, and (2) compression, which reduces the size of encrypted gradients before blockchain storage. These techniques aim to improve efficiency without compromising quality. Continued research seeks to enhance scalability and make this approach more practical.

\section{Future Lines of Research}
\label{sec:futureworks}



The field of blockchain-based federated learning presents several promising avenues for future research. One such area is investigating incentive systems that can encourage data providers to participate in the federated learning process within blockchain networks. Incentive systems can help address challenges such as the lack of motivation for data providers to contribute to federated learning or the potential for free-riding on the contributions of others. The design and implementation of incentive mechanisms that provide adequate rewards to data providers without compromising the security and privacy of the system is a crucial research question.

Another promising direction for future research in blockchain-based federated learning is the development of privacy-preserving techniques. Although current systems use secure aggregation to encrypt model updates before transmitting them to the server, this approach may not be sufficient to protect sensitive data. Therefore, exploring novel techniques for data privacy protection in federated learning is necessary. Such techniques could include encryption methods such as homomorphic encryption, differential privacy, or secure multi-party computation. These techniques ensure that the data remains private, even from the server or other nodes, while allowing for effective aggregation and model updates.

Smart contracts, which are significant programs that run on the blockchain, can also play a significant role in blockchain-based federated learning. Smart contracts can automate and enforce the rules governing the training process, including data privacy and security protocols and incentives for participating nodes. Using smart contracts could enhance the transparency and fairness of the system, allowing for more efficient and secure federated learning.

Moreover, integrating machine learning techniques such as transfer learning and meta-learning can enhance the efficiency and effectiveness of the federated learning system. Transfer learning is a machine learning technique that allows the transfer of knowledge from one task to another. On the other hand, meta-learning is a learning approach that utilizes prior knowledge to achieve faster and more accurate learning. Incorporating these techniques into the federated learning system can reduce the required training data and improve the models' accuracy.

\section{Conclusion}
\label{sec:Conclusion}
Blockchain-based Federated Learning (FL) is an emerging approach that has garnered significant interest in improving the privacy and security of machine learning models. The decentralized and immutable nature of blockchain technology has the potential to replace traditional centralized methods, enhancing the privacy and efficiency of FL. Blockchain technology can protect against data breaches and malicious actors while enabling multiple parties to train models collaboratively without sharing their data. Moreover, data distribution among multiple parties without a centralized server can further reduce the risks of data breaches and enhance data privacy. The use of blockchain technology also ensures that data is securely stored and accurately tracked, enabling efficient and trustworthy collaborative learning. The integration of blockchain and FL holds immense potential in advancing the field of machine learning and improving its practical applications in various industries.

\bibliography{refs}{}
\bibliographystyle{plain}

\end{document}